# Design of an Alarm System for Isfahan's Ozone Level based on Artificial Intelligence Predictor Models

Ehsan Lotfi
Department of Computer Engineering, Torbat-e-Jam Branch, Islamic Azad University, Torbat-e-Jam, Iran
Email: esilotf@gmail.com

**Abstract-** The ozone level prediction is an important task of air quality agencies of modern cities. In this paper, we design an ozone level alarm system (OLP) for Isfahan city and test it through the real word data from 1/1/2000 to 7/6/2011. We propose a computer based system with three inputs and single output. The inputs include three sensors of solar ultraviolet (UV), total solar radiation (TSR) and total ozone (O3). And the output of the system is the predicted O3 of the next day and the alarm massages. A developed artificial intelligence (AI) algorithm is applied to determine the output, based on the inputs variables. For this issue, AI models, including supervised brain emotional learning (BEL), adaptive neuro-fuzzy inference system (ANFIS) and artificial neural networks (ANNs), are compared in order to find the best model. The simulation of the proposed system shows that it can be used successfully in prediction of major cities' ozone level.
**Keywords** – ozone predictor, artificial intelligence, UV

## 1. Introduction

Prediction of ozone level is an important factor for modern urban management. A proper predictor can help the air quality agencies for decision-making and announcing the alarms. The ozone level can be considered as a chaotic time series. In the literature, there are two approaches for forecasting ozone level, including 1) non-linear differential equation based methods and 2) neural based methods. Mathematical models include kalman filter, ARMA, ARCH and GARCH etc. and Artificial intelligence (AI) methods consist of learning based methods such as multilayer neural network (ANN), Adaptive Neuro-Fuzzy Inference Systems (ANFIS) [38-44] and the brain emotional learning (BEL). AI methods can learn the behavior of the data by observing the previous values, and can predict the future values. Firstly, this paper aims to design an alarm system of ozone level base on AI models such as BEL. Secondly, here we compare the AI's methods including BEL, ANFIS and ANN in the Isfahan's ozone level prediction task in order to find the best predictor. Isfahan is a modern city of Iran, and the ozone level prediction is very important for its urban management. So we consider the ozone level dataset of Isfahan city as a case study to compare the AI's models.

The organization of the paper is as follows: The proposed system is described in Section 2. The Comparative results are presented in Section 3. Finally the conclusions are made in Section 4.

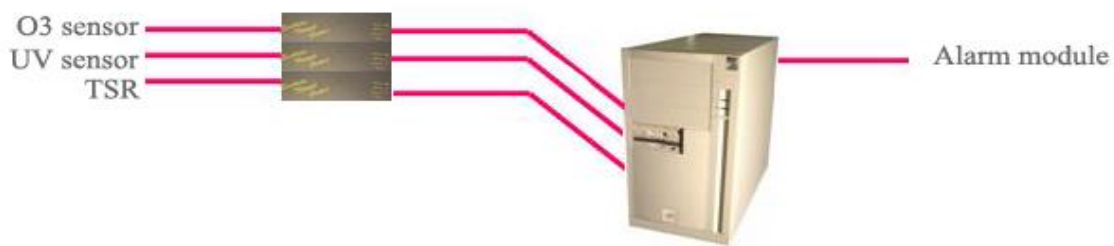

Fig. 1. General view of proposed system

## 2. Proposed system

The geomagnetic studies [47] has shown that solar ultraviolet (UV), total solar radiation (TSR) and current total ozone (O3) can affect the next level of O3. So our system applies them to make the alarms. The general view of proposed system is presented in Fig. 1. The proposed system is very simple. There are three sensors taht fed a program based on AI methods run on the computer. And the computer fed an alarm module or any display or massaging module. In order to find the best AI model, we compare three well know models of this field, including BEL, ANFIS and ANN.

**2.1. BEL**

The BEL algorithm comes from a neurophysiological aspect of the brain, shows to be a novel proper learning algorithm and is an appropriate online predictor for complex systems. The supervised BEL approach (Lotfi and Akbarzadeh 2013; 2014a; 2014b; 2014c) [10] can learn pattern-target samples and can apply well to prediction problems. Furthermore, BEL profits a decay mechanism for Amygdala and a monotonic learning rule. It is a novel recognizer artifact with very simple structure that is motivated by neurophysiological knowledge of the limbic system of the brain [13-20]. Supervised BEL can be used in pattern recognition, classification, prediction and the fitting problems. The supervised BEL is utilized here to predict the ozone level of Isfahan city, where the low value of ozone is very dangerous.

Fig. 2 shows supervised BEL. It's a feed forward neural network that is based on amygdala-orbitofrontal interacts. The functionality of amygdale is considered as a threshold logic unit. This structure can be distributed as a multi output structure and is considered the target values in learning process. This structure is model free and can be used in multi input multi output classification and prediction problems. The learning algorithm of the model presented in Fig.2, has been described in (Lotfi and Akbarzadeh 2012, 2013). The learning algorithm includes two rules which are as follows:

$$v_j^{k+1} = (1-\gamma)v_j^k + \alpha \max(t^k - E_a^k, 0) p_j^k \tag{1}$$

$$w_j^{k+1} = w_j^k + \beta(E^k - t^k) p_j^k \tag{2}$$

where $k$ is learning step, $\alpha$ and $\beta$ are learning rates and $\gamma$ is decay rate in amygdala learning rule, where the $t^k - E_a^k$ is calculated error. This model has been used in various applications [1-12] and [20-34] and is developed here to predict the ozone level problem.

**2.2 ANFIS**

ANFIS [45] combines the fuzzy approaches and artificial neural networks (ANNs) in order to use the advantages of them. In ANFIS model, the fuzzy rules and membership functions are generated automatically by a neural network learning algorithm. Fig. 3 shows the architecture of Sugeno-type ANFIS. This architecture includes the following five layers; fuzzifier, production, normalized, defuzzy, and output layer.

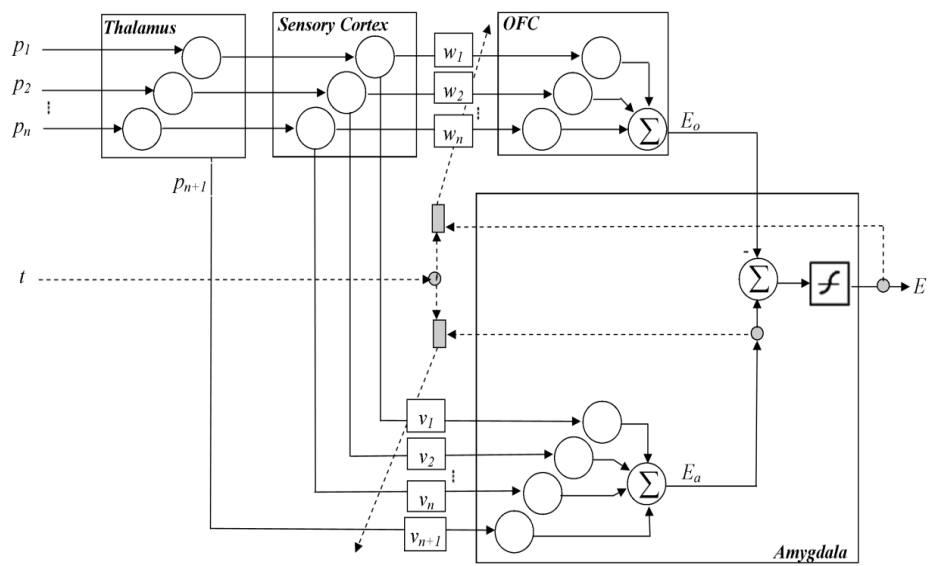

Fig. 2 Three-inputs single output architecture of supervised BEL (Lotfi 2013)

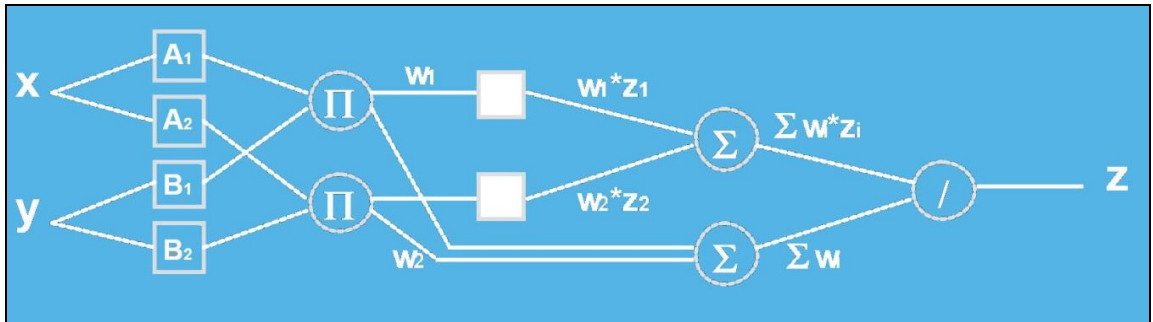

Fig. 3 Architecture of ANFIS

In this sample ANFIS architecture presented in Fig. 3, the learning parameters including *w*1, *w*2, *z*1, *z*2 and *A*1, *A*2, *B*1, *B*2 should be adjusted. This adjustment is performed by using input-target examples (Jang 1997) [45].

## 2.3 ANN

ANN is inspired by physiological workings of the brain. ANN resembles the actual networks of neural cells (neurons) in the brain. ANN have particular characteristics such as the ability to learn and generalize. ANN learns through the error back propagation algorithm. According to this algorithm, the error of output units is propagated back to adjust the connecting weights within the network. The 2-2-1 architecture of ANN is presented in Fig. 4. It's a two input-single output architecture with two neurons in hidden layer. This architecture has single hidden layer with two hidden neurons.. After learning, the inputs and output of this model involve the following equation,

$$\hat{z}_t = f_{net}(x_t, y_t) \tag{3}$$

Where $\hat{z}_t$ is predicted *z* value at time *t*.

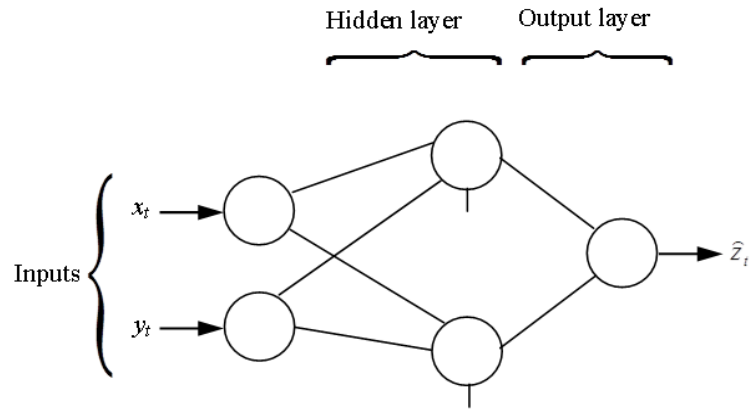

Fig. 4 Architecture 2-2-1 of ANN

## 3. Simulation results

To assess the model, a dataset is obtained from Isfahan's research institutes. The dataset is accessible from www.bitools.ir. The dataset includes the daily ozone level of Isfahan from 1/1/2000 to 7/6/2011. Fig 4 shows the first 500 values of the Isfahan's ozone level time series. The dataset is divided into the three parts; the first is the training set that includes 70% of data selected randomly. And the second is the validating set that is 15% data and the thirds is testing set where the prediction results are evaluated and it's 15% of the data. Fig. 6 illustrates the comparative results of prediction obtained from BEL, ANN and ANFIS. According to the results the ANFIS shows higher correlation that BEL and ANN. Figs. 7 and 8 show the target and predicted level obtained from BEL and ANN respectively.

Figs. 7 and 8 are the curve of predicted level versus observation from 1/1/2000 to 7/6/2011. As illustrated in the figures, the best COR = 0.80234 is obtained from the BEL model. In these experiments, the models have applied the previous 4 values of each point of ozone level. So the size of input patterns is 4.

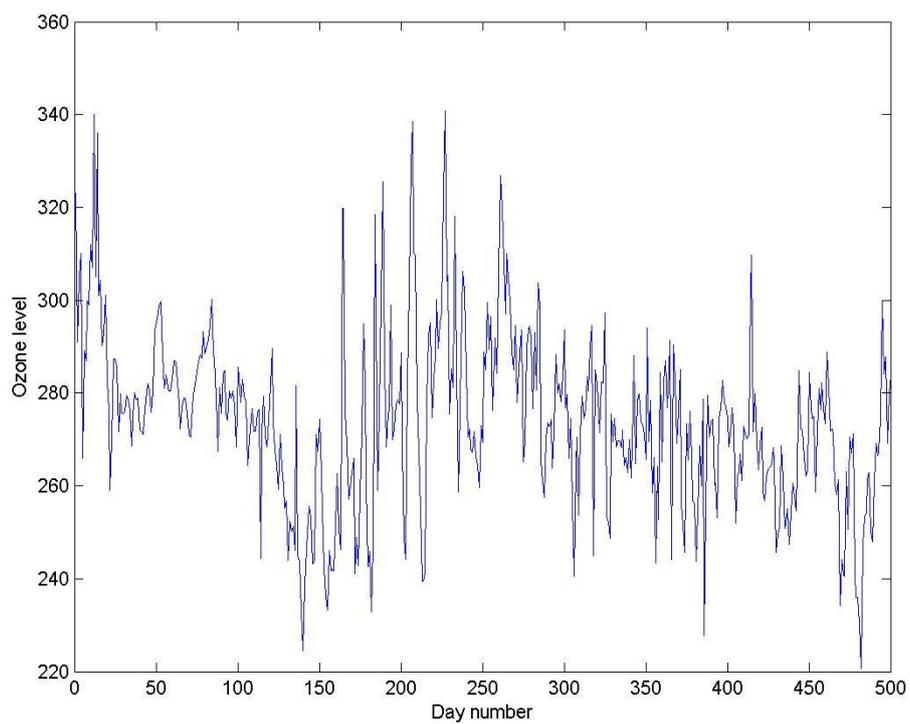

Fig. 5. The ozone level of the first 500 days in Isfahan's dataset.

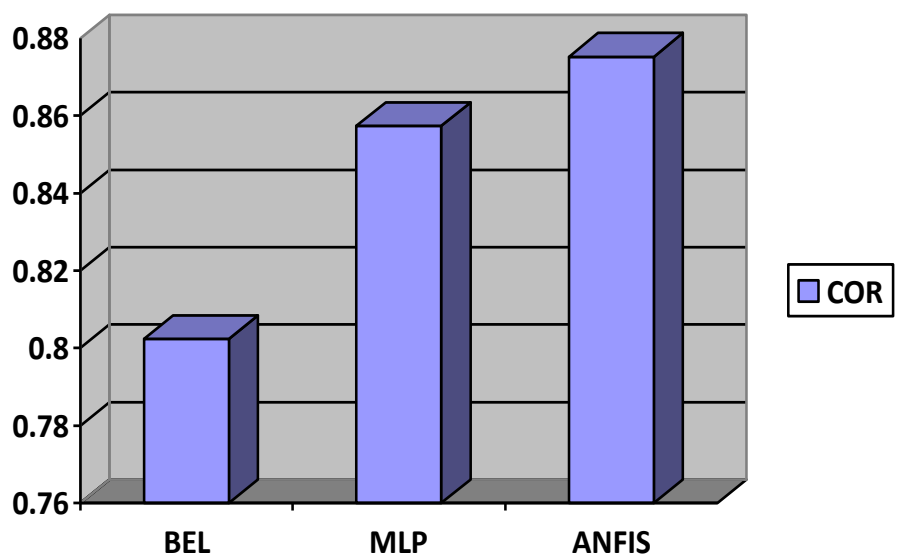

Fig. 6 The correlation comparison between BEL, ANN and ANFIS in the ozone level prediction.

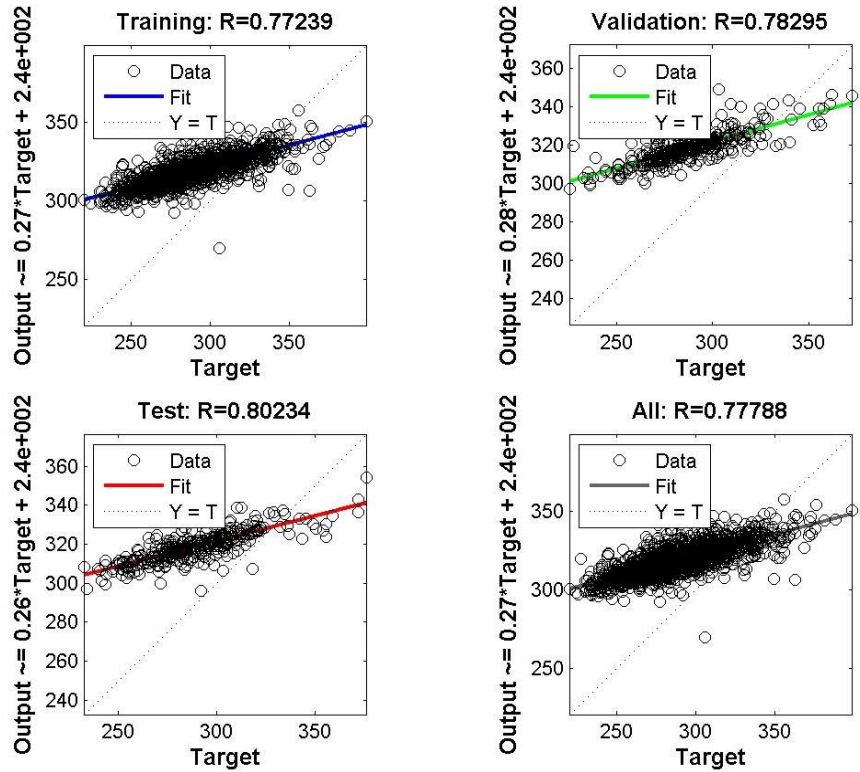

Fig. 7 the correlation of the results obtained from BEL in the training set, test set and validation set.

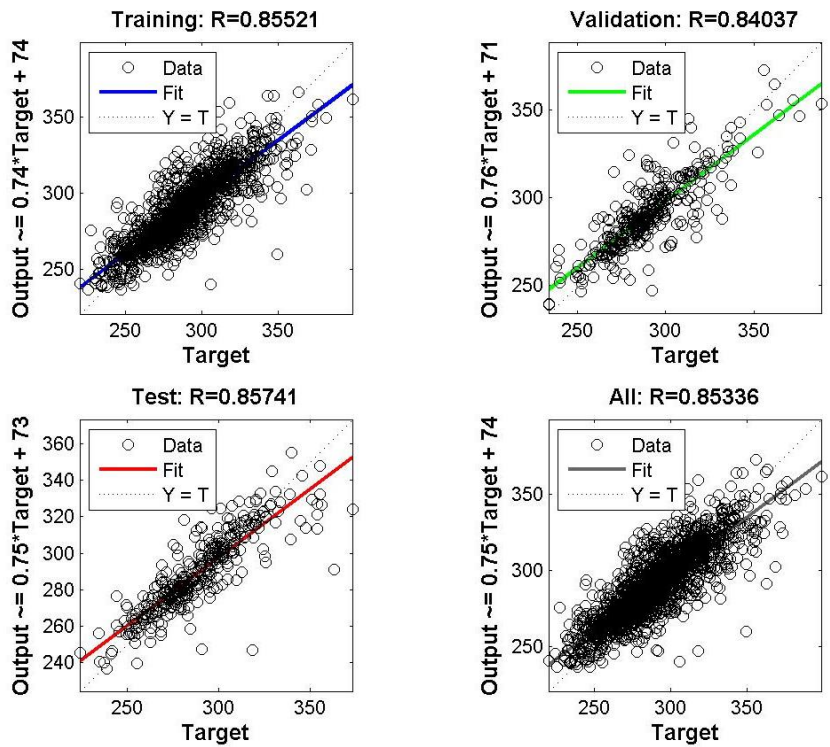

Fig. 8 the correlation of the results obtained from ANN in the training set, test set and validation set.

Thus in the system presented in Fig. 1, ANFIS algorithm is an optimum solution for ozone prediction. The final step of design is associated with determination of a logical threshold on predicted O3 to set the alarm massages. This threshold should be determined manually and by expert, and system manager.

4. Conclusions

In this paper, firstly we propose a novel alarm system for ozone level. Secondly we develop the BEL algorithm for ozone level prediction task and compare the results of BEL, ANFIS and ANN in order to find the best AI model in the system. In the numerical studies, BEL, ANFIS and ANN is utilized to predict the Isfahan's ozone level. According to the simulations, ANFIS shows higher correlation in the prediction task. ANFIS associated with three input sensors of O3, UV and TSR and with a determined threshold can be used as an alarm module for major cities.